\documentclass[journal]{IEEEtran}
\ifCLASSINFOpdf
\else
\fi
\usepackage{amssymb}
\usepackage{caption}
\usepackage{subcaption}
\usepackage{graphicx} 
\usepackage{pifont}
\usepackage{array}
\usepackage{graphicx}
\usepackage{cite}

\usepackage{comment}
\usepackage{amsmath,amssymb} % define this before the line numbering.
\usepackage{color}
\usepackage{multirow}
\usepackage{multicol}
\usepackage{booktabs}
\makeatletter
\let\NAT@parse\undefined
\makeatother
\usepackage{hyperref}  %hyperref still needs to be put at the end!
\usepackage{colortbl}
\usepackage{color}
\usepackage[table]{xcolor}
\newcommand{\mypm}[1]{\color{gray}{\tiny{#1}}}
\definecolor{lightgray}{rgb}{0.9, 0.9, 0.9}
\begin{document}
% \linenumbers
\title{A Survivor in the Era of Large-Scale Pretraining: An Empirical Study of One-Stage Referring Expression Comprehension}

\author{Gen~Luo,
        Yiyi~Zhou*,~\IEEEmembership{Member,~IEEE,}
        Jiamu Sun, 
        Xiaoshuai Sun,~\IEEEmembership{Member,~IEEE,} 
        Rongrong Ji,~\IEEEmembership{Senior~Member,~IEEE }% <-this % stops a space

        \thanks{G. Luo, Y. Zhou, J. Sun,  X. Sun and R. Ji are with the Key Laboratory of Multimedia Trusted Perception and Efficient Computing, Ministry of Education of China, Xiamen University, 361005, P.R. China. G. Luo and R. Ji are also with the Peng Cheng Laboratory, Shenzhen 518000, China. (e-mail: luogen@stu.xmu.edu.cn, zhouyiyi@xmu.edu.cn, sunjiamu@stu.xmu.edu.cn, xssun@xmu.edu.cn, rrji@xmu.edu.cn).}% <-this % stops a space 
 
        \thanks{*Corresponding Author: Yiyi~Zhou (E-mail: zhouyiyi@xmu.edu.cn)}
%\thanks{Yiyi Zhou was with the Department
%of Electrical and Computer Engineering, Georgia Institute of Technology, Atlanta,
%GA, 30332 USA e-mail: zhouyiyi@xmu.cn.}% <-this % stops a space
}


\markboth{Journal of \LaTeX\ Class Files,~Vol.~14, No.~8, August~2015}%
{Shell \MakeLowercase{\textit{et al.}}: Bare Demo of IEEEtran.cls for IEEE Journals}



\maketitle

% Use \bibliography{yourbibfile} instead or the References section will not appear in your paper 
\begin{abstract}
	One-stage Referring Expression Comprehension (REC) is a task that requires accurate alignment between text descriptions and visual content. In recent years,  numerous efforts have been devoted to cross-modal learning for REC, while the influence of other factors in this task still lacks a systematic study.  To fill this gap, we conduct an empirical study in this paper.  Concretely, we  ablate  42  candidate designs/settings based on a  common REC framework,  and these candidates  cover  the entire process of one-stage REC from network design to model training.  Afterwards, we  conduct over 100 experimental trials on three  REC  benchmark datasets. The extensive experimental results     reveal  the key factors that affect REC performance in addition to multi-modal fusion,  {e.g.},  multi-scale features and data augmentation.  Based on these findings, we further propose a simple yet strong model called SimREC, which achieves new state-of-the-art performance on  these benchmarks.   In addition to these progresses, we also  find  that with much less training  overhead and  parameters, SimREC can   achieve better performance than a set of large-scale pre-trained  models, e.g., UNITER and VILLA, portraying the special role  of REC  in existing V\&L research\footnote{Source codes have been  released at: {https://github.com/luogen1996/SimREC}.}.
\end{abstract}

%%%%%%%%% BODY TEXT
\section{Introduction}
\label{sec:intro}
Referring expression comprehension~\cite{zhou2021real,MCN,MATT:,sun2021iterative,wang2019neighbourhood}, also known as \textit{visual grounding}~\cite{liu2019learning,yang2019fast,ReSC,deng2021transvg}, is a task of locating the target instance in an image based on the natural language expression.  As an vision and language (V\&L) task, REC is not limited to a fixed set of object categories and can theoretically perform any open-ended detection  according to    text descriptions~\cite{zhou2021real}. These appealing properties enable REC to garner widespread  attention from the communities of both computer vision (CV) and vision and language (V\&L)~\cite{zhou2021real,MCN,MATT:,sun2021iterative}.

\begin{figure}[t]
	\centering
	\includegraphics[width=1\columnwidth]{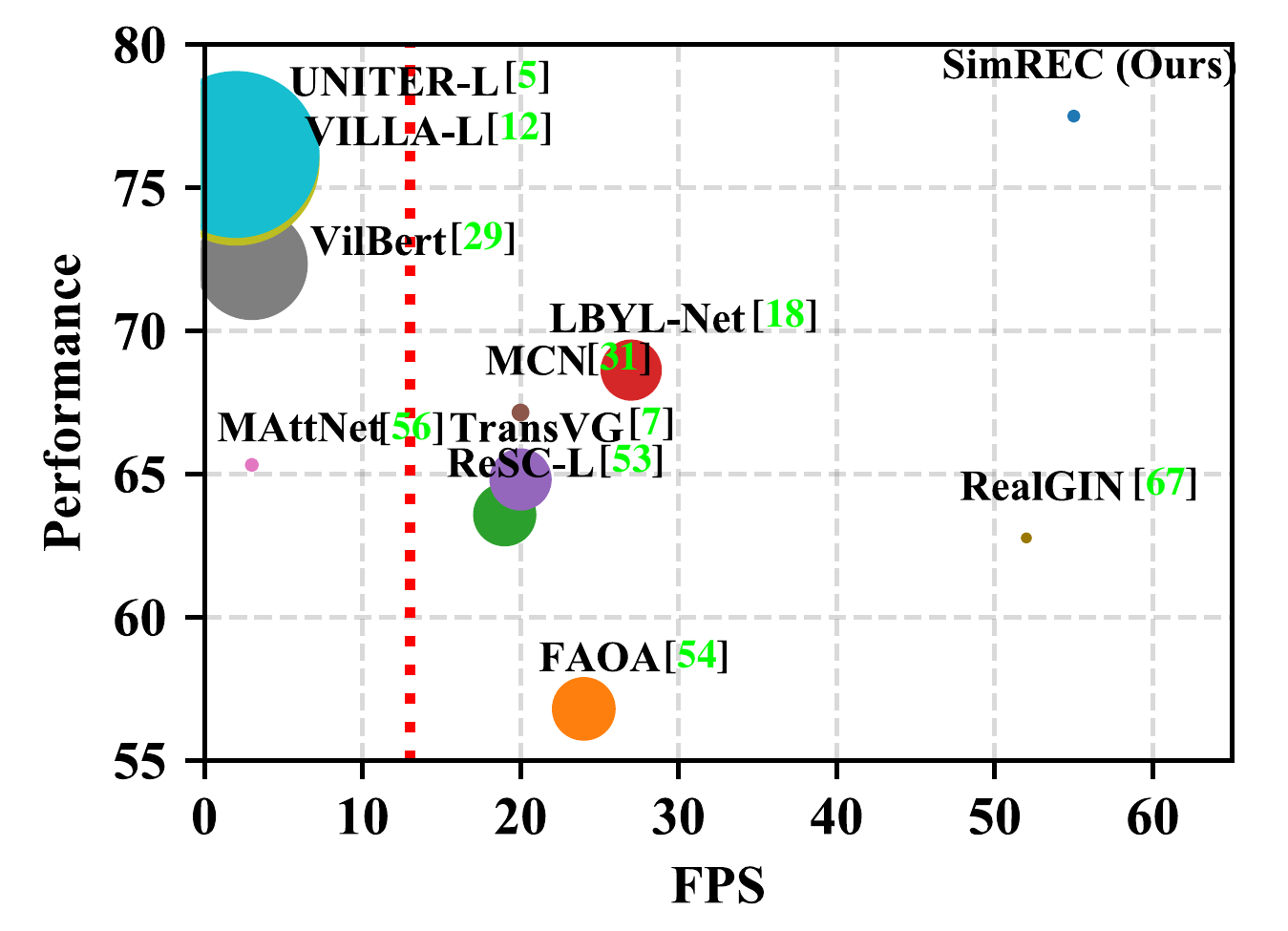}
	%	\vspace{-7mm}
	% Reduce the figure size so that it is slightly narrower than the column. Don't use precise values for figure width.This setup will avoid overfull boxes. 
	\vspace{-2em}
	\caption{Comparison of performance and inference speed between SimREC and other REC methods on the \textit{val} set of RefCOCO+~\cite{REFCOCO}.   The point size  corresponds to the number of parameters.   FPS denotes frames per second.  Methods on the left side of the red line are two-stage, while the rest are one-stage.  } 
	\vspace{-5mm}
	\label{fig1-1}
\end{figure}
Due to the obvious advantage in efficiency, one-stage modeling has recently become the main research focus in REC~\cite{zhou2021real,MCN,yang2019fast,ReSC,deng2021transvg}. %我们的方法先引用
Compared with the two-stage methods~\cite{MATT:,wang2019neighbourhood,bajaj2019g3raphground,liu2019improving,liu2019learning}, one-stage models can omit the generation of region proposals and directly output the bounding box of the referent without complex image-text ranking, thereby improving the inference speed by an order of  magnitude, as shown in Fig.~\ref{fig1-1}. However, one-stage models  often have worse performance  than the two-stage ones~\cite{zhou2021real,yang2019fast,MCN}. Practitioners mainly attribute this shortcoming to the lack of enough multi-modal reasoning ability~\cite{ReSC,deng2021transvg,zhou2021real,huang2021look}. To this end, recent endeavors put numerous efforts into the design of the multi-modal  networks for  one-stage REC, which have  achieved notable success.

Despite these advances, the existing literature still lacks a   systematic   study to  gain insight into   one-stage REC. Differing from other V\&L tasks, one-stage REC not only needs to realize the joint understanding of two modalities, but also requires  very accurate language-vision alignments, \emph{i.e.}, locating the referents with bounding boxes. Therefore, in addition to multi-modal fusion, other important factors also need to be further investigated, such as visual backbone, detection head and training paradigm.  
In this paper, we are committed to filling this research gap.  We aim to figure out   \textit{what also matter in one-stage REC}, and\textit{ how  they  affect the  REC performance}. Meanwhile, we also question  that  

\textit{ whether REC, as an important V\&L task,  can be a survivor in the era dominated by large-scale BERT-style pre-trained models}?

\begin{figure*}[t]
	\centering
	\includegraphics[width=1.9\columnwidth]{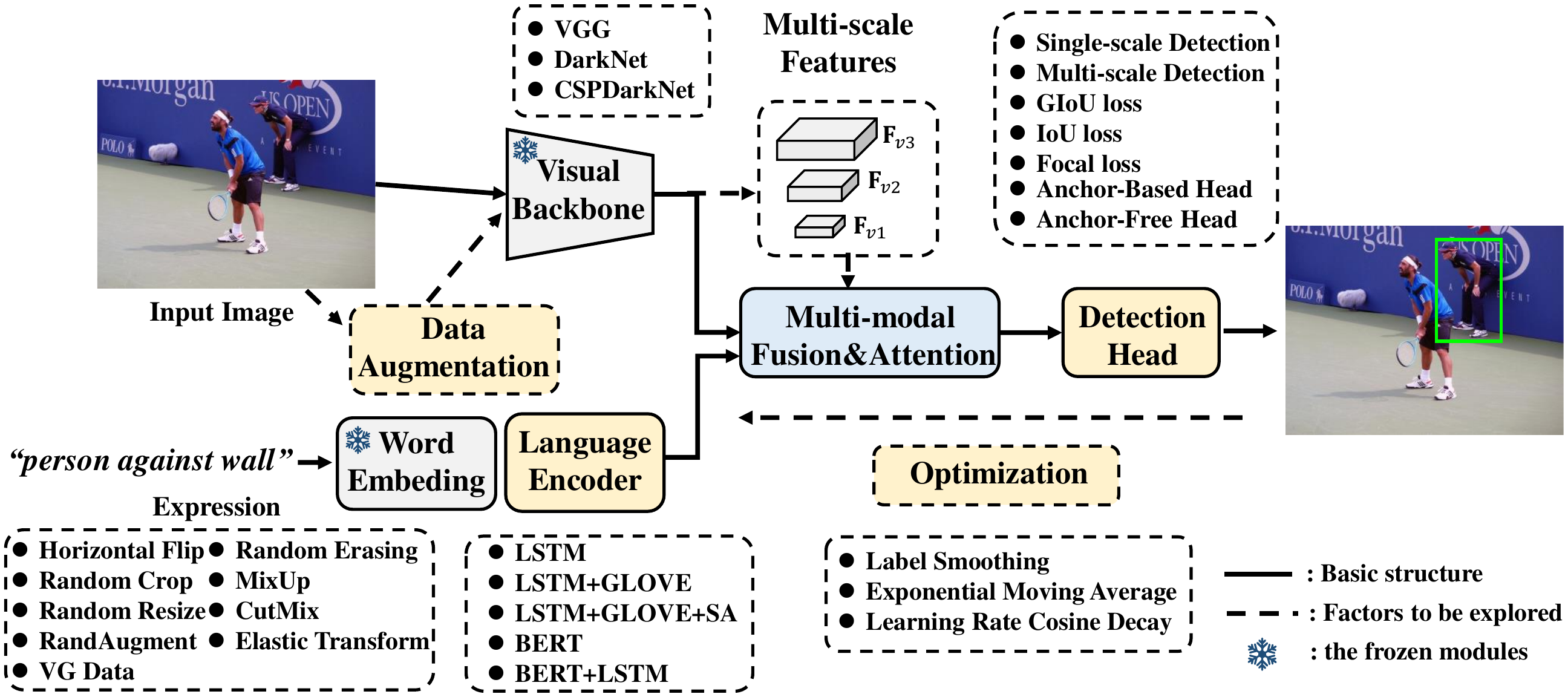}
	
	% Reduce the figure size so that it is slightly narrower than the column. Don't use precise values for figure width.This setup will avoid overfull boxes. 
	\caption{The model framework and candidate designs/settings. 
		The solid lines depict the basic structure.  
		The dashed parts list candidate designs for the empirical study,  which are used  to examine the effects of   different  components in one-stage REC. {The visual backbone and word embeddings are frozen during training.} }
	\label{framework}
	\vspace{-1em}
\end{figure*}

%另起一段
To approach these targets, we first build  a common  one-stage REC network  framework, as shown  Fig.~\ref{framework}.  Based on this framework, we conduct  an empirical research  that ablates 42 candidate designs / settings, which cover the most components of one-stage REC, such as visual backbone, language encoder, detection head and training paradigm \emph{et al.}   Afterwards, we conduct over 100   experimental trials  on three benchmark datasets, namely RefCOCO~\cite{REFCOCO}, RefCOCO+~\cite{REFCOCO} and RefCOCOg~\cite{nagaraja2016modeling}, to examine the effects of these candidates. Lastly, these  results constitute the main analysis data of this research.

%\begin{figure}[t]
%	\centering
%	\includegraphics[width=1\columnwidth]{fig1-2.pdf}
%	%	\vspace{-7mm}
%	% Reduce the figure size so that it is slightly narrower than the column. Don't use precise values for figure width.This setup will avoid overfull boxes. 
%	\caption{Statistical results of the impact of  components in one-stage REC on the \emph{val} set of RefCOCO+. } 
%	%	\vspace{-3mm}
%	\label{fig-21}
%\end{figure}

This study yields significant findings for one-stage REC. Above all,  it reflects the key factors that affect performance. Visual backbone and multi-scale visual features are  critical for REC performance. And data augmentation, including  image augmentation methods~\cite{zhong2020random,cubuk2020randaugment,yun2019cutmix,devries2017improved} and additional data~\cite{krishna2016visual}, is also very beneficial. 
Meanwhile, we obtain some findings  against the conventional impressions about REC. For example, REC is less affected by language bias, which is however very common in other V\&L tasks like visual question answering~\cite{VQA0,VQA1,zhou2019plenty}. 
In addition, we also notice that although some candidate designs obtain the expected performance gains, their practical use is still different to that in other tasks, \emph{e.g.}, augmentation methods~\cite{zhong2020random,cubuk2020randaugment,yun2019cutmix,devries2017improved}. 
Detailed discussions are given in the experimental section.

By adopting these empirical findings, we further propose a simple yet strong model called SimREC, which    outperforms most existing one-stage and two-stage methods in both accuracy and inference speed, as shown in Fig.~\ref{fig1-1}. More importantly, with less training overhead and parameters, SimREC can achieve better performance  than  a bunch of  large-scale pre-trained models~\cite{lu2019vilbert,chen2019uniter,gan2020large,yu2020ernie}, \emph{e.g.,} VILLA-Large~\cite{gan2020large}, on all benchmark datasets. We believe this result can be very enlightening for the existing V\&L research, where the most tasks are dominated by  the  expensive large-scale models. 

Overall, the contributions of this paper are three-fold:

\begin{itemize}
	\item We present the first systematic study for one-stage REC, yielding several key factors in addition to multi-modal fusion. 
	\item We propose a strong and simple model called SimREC, which outperforms a set of REC methods and large-scale pre-trained models in both performance and efficiency.
	\item We build a comprehensive and easy-to-use codebase\footnote{  {Our codebase is  lightweight, extendable and general, and  new REC models can be developed with minimal efforts.  It also supports large-scale pre-training and multi-task learning of REC and RES on six common datasets, which is valuable for existing RES and REC researches.}} based on the content of this paper, which can greatly promote the development of one-stage REC. 	
\end{itemize}

\section{Related Work}
Referring Expression Comprehension (REC)~\cite{hu2017modeling,hu2016natural,liu2017referring,luo2017comprehension-guided,yu2016modeling,yu2017a,zhang2017discriminative,yu2018mattnet:,wang2019neighbourhood,MCN,ReSC,zhou2021real,bu2022scene,sun2022proposal,qiao2020referring,luo2022towards,luo2021towards,zhou2019plenty}  is a  task to locate the  referent based on a  natural language  expression.  Early works~\cite{hu2017modeling,liu2017referring,luo2017comprehension-guided,yu2016modeling,zhang2017discriminative,yu2018mattnet:,wang2019neighbourhood,zhou2021trar}   often follow a two-stage pipeline.  Concretely,  two-stage models first detect the salient regions of an image,  and then regard REC task as region-expression ranking problem. Despite the great success, these two-stage methods  have obvious defects in model efficiency and generalization~\cite{yang2019fast,ZSGNet}.

To this end, one-stage REC has recently become a popular research direction~\cite{yang2019fast,MCN,ReSC,ZSGNet,zhou2021real,sun2021iterative,huang2021look,zhu2022seqtr,liao2022progressive,zhao2022word2pix,ye2022shifting}.  
By omitting the steps of region detection and image-text matching in multi-stage modeling, one-stage models greatly reduce the inference time, \emph{e.g.,} 28.6 images per second, as reported in \cite{zhou2021real}. However, one-stage models often have worse REC performance compared with the two-stage methods, like MattNet~\cite{MATT:}, mainly due to the limited reasoning ability.   In this case, recent advances  are dedicated  to improving the reasoning ability of one-stage REC  and propose various novel multi-modal networks for one-stage REC~\cite{ReSC,deng2021transvg,zhou2021real,huang2021look,luo2022towards,zhou2021trar}.  {In particular, ReSC~\cite{ReSC} proposes a multi-round reasoning module to handle long and complex expressions.  LBYL~\cite{huang2021look}  proposes a novel landmark feature convolution  to better align the vision and language features.  TransVG~\cite{deng2021transvg} improves  multi-modal fusions via stacking multi-layer transformer blocks~\cite{vaswani2017attention}. More recently, researchers have formulated REC as a sequence prediction task and proposed a set of novel sequence-to-sequence frameworks~\cite{zhu2022seqtr,liu2023polyformer}. For example, SeqTR~\cite{zhu2022seqtr} represents the bounding box of the referent with a sequence of discrete coordinate tokens, which are predicted via a Transformer architecture.  In addition to the design of the network architecture,  numerous efforts  are devoted into multi-task learning~\cite{MCN,zhu2022seqtr,liu2023polyformer,yan2023universal,wang2022ofa,zhang2023one}, and greatly improve the upper bound of REC models.      Despite the effectiveness,  except the well studied multi-modal fusion, there still lacks a systematic study to examine the key  components in one-stage REC.}

{
Driven by the  progresses of large-scale V\&L pre-training, the new SOTA performance of REC is  achieved by recent large-scale pre-training models \cite{lu2019vilbert,chen2019uniter,gan2020large,yu2020ernie,kamath2021mdetr}. Specifically, these models are pre-trained with millions of image-text examples and  excessive  parameters, thereby dominating various downstream VL tasks, \emph{e.g.,} VQA~\cite{VQA0} and multi-modal retrieval~\cite{MSCOCO}.   In particular,  common pre-training objectives of early  VL pre-training  are masked language modeling~\cite{chen2019uniter,devlin2018bert:} and image-text matching~\cite{lu2019vilbert,chen2019uniter}.  When transferring to REC task, most of these models  follow the two-stage paradigm and regard this task as a  text and image region matching problem~\cite{lu2019vilbert,chen2019uniter,gan2020large,yu2020ernie}.
Nevertheless, these BERT-style pre-trained models are often defective in their expensive training costs and slow inference speed. Meanwhile, their pre-training objectives are also inefficient for the adaptation to REC task.   
In this paper, we   are  keen to   find out whether a simple and efficient one-stage model can  outperform   these expensive pre-trained models.  }

\section{The Framework and Ablation Designs}
To accomplish the empirical study, we first build a common one-stage REC framework, as depicted in   Fig.~\ref{framework}.
Then, we prepare a set of  ablation designs and settings to examine the  impact of different components.
\subsection{Framework}

Specifically, given an image $I$ and an expression $E$,  an one-stage REC model often  uses a visual backbone and a language encoder to extract the visual and text features, denoted as $\textbf{F}_v \in \mathbb{R}^{h \times w \times d}$ and $\textbf{F}_t \in \mathbb{R}^{ l \times d}$, respectively.  Here, $h\times w$ denotes the resolution of visual feature maps, and $l$ is the length of the expression. Here,  the text features $\textbf{F}_t$ are  attentively pooled~~\cite{MCAN} to a global vector $f_t \in \mathbb{R}^{d}$.
Afterwards, a simple multi-modal fusion module is deployed to obtain the joint representations of two modalities, denoted as $\textbf{F}_m  \in \mathbb{R}^{h \times w \times d}$, which is obtained by:
\begin{equation}
f_m= \sigma(f_v\textbf{W}_v) \odot \sigma(f_t\textbf{W}_t).
\label{eq1}
\end{equation}
Here, $f_m$ and $f_v$ are the feature vectors of $\textbf{F}_m$ and $\textbf{F}_v$, respectively, and $\sigma$ denotes the   ReLU~\cite{RELU}  activation function. To keep a certain reasoning ability,   we also apply  an attention  unit called GARAN~\cite{zhou2021real}   after the multi-modal fusion.

Lastly, a regression layer is deployed to predict the bounding box of the referent in the image. 
Specifically,  for each grid $(g_x, g_y)$, SimREC predicts raw coordinates of the bounding box  ${\{ {t_x},{t_y},{t_w},{t_h}\}}$ and its corresponding confidence score $c$. Afterwards, the final bounding box ${b=\{ {b_x},{b_y},{b_w},{b_h} \}}$  is calculated based on the pre-defined anchor box $(p_w, p_h)$ by 
\begin{equation}
\begin{aligned}
&b_x=\text{sigmoid}({t_x})+g_x, \\ &b_y=\text{sigmoid}({t_y})+g_y,\\
&b_w=p_w^{e^{{t_w}}}, b_h=p_h^{e^{{t_h}}}.
\end{aligned}
\label{EQ1}
\end{equation}
Given the ground-truth  box ${b'=\{ {b_x'},{b_y'},{b_w'},{b_h'} \}}$ and the  ground-truth confidence $c'$, the loss function of SimREC  is  defined as:
\begin{equation}
\small
\begin{aligned}
&l(b_i,b_i',c_i,c_i')=\sum_{i=1}^{h \times w \times n }c'_i*l_{box}(b'_i,b_i)+l_{conf}(c'_i,c_i)
%	&l_{reg}=\sum_{i=1}^{h \times w }{(b'-b_i)^2}*c_i', \\
%&l_{conf}=\sum_{i=1}^{h \times w}{c_i' \text{log}(c_i)+(1-c_i')\text{log}(1-c_i)},\\
%&l=	l_{reg}+l_{conf}.
\end{aligned}
\end{equation}
Here, $n$ is the number of pre-defined anchor boxes. $l_{box}$  and $l_{conf}$ denote the regression loss and  the
binary prediction loss of the confidence score, respectively. During the test stage, SimREC will select the bounding  box with the highest confidence score as the final prediction. 

{
Although the framework is simple, it includes the main design patterns of most existing one-stage REC models, \textit{i.e.,}  image encoder, language encoder, fusion branch and detection head.   Meanwhile, a simple framework can be used to better examine other key factors for one-stage REC in addition to multi-modal fusion. With this simple framework, we can better ablate the key factors in one-stage REC.  }

% \usepackage{multirow}

\subsection{Ablation Designs}

\textbf{Visual Backbone.} We use  DarkNet~\cite{redmon2018yolov3:} as the default visual backbone. VGG~\cite{simonyan2015very} and the recently proposed CSPDarkNet~\cite{wang2020cspnet} are  the candidate choices. Meanwhile, we also test the effects of different image resolutions.

\textbf{Language Encoder.} The default language model   is an LSTM network~\cite{LSTM} with GLOVE embeddings~\cite{pennington2014glove}. During experiments, we will enhance the ability of language encoder by adding advance designs like self-attention~\cite{vaswani2017attention} and BERT embedding~\cite{devlin2018bert:}.

\textbf{Multi-scale Features and Detection}.  We also experiment  the model  with multi-scale features and multi-scale detection setting. In terms of multi-scale features, we  use the last $k$ scale feature maps of the visual backbone as our visual features, denoted as $\{\textbf{F}_{vi}\}_{i=1}^{k}$.  The features of each scale are  fused with the text ones by the multi-scale fusion scheme  proposed in \cite{MCN}, denoted as $\{\textbf{F}_{mi}\}_{i=1}^{k}$.  Only the last scale feature maps are used, \emph{i.e.}, $\textbf{F}_{mk}$.  In terms of multi-scale detection, each scale feature in $\{\textbf{F}_{mi}\}_{i=1}^{k}$ will use a corresponding detection head to predict bounding boxes.

\textbf{Detection Paradigms.}
In addition to the anchor-based detection head~\cite{redmon2018yolov3:}, we also  experiment the recently popular anchor-free detection~\cite{ge2021yolox}. In anchor-free detection, the model directly predict the width and height of  bounding boxes for each grid.
The objectives we test include \textit{IoU} loss~\cite{rezatofighi2019generalized}, \textit{GIoU} loss~\cite{rezatofighi2019generalized}, \textit{Smooth-L1} loss~\cite{ren2017faster} and \textit{Focal} loss~\cite{lin2017focal}.

\textbf{Data Augmentation.} We adopt additional data, \emph{i.e.}, Visual Genome~\cite{krishna2016visual}, and image augmentation methods~\cite{zhong2020random,yun2019cutmix,cubuk2020randaugment,devries2017improved,zhang2017mixup}  for data augmentation. In terms of image augmentation, we try   \textit{ElatiscTransform}, \textit{CutMix}~\cite{yun2019cutmix}, \textit{Random Resize}, \textit{RandAugment}~\cite{cubuk2020randaugment} and \textit{MixUp}~\cite{zhang2017mixup}. Their detailed modification for REC are described in our appendix. 

\textbf{Optimizations.} To regularize and accelerate the model training, we  also examine  some training  techniques,  \emph{e.g.},  label smoothing~\cite{he2019bag}, exponential moving average (EMA)~\cite{tarvainen2017mean} and   cosine decay learning schedule~\cite{he2019bag}.

\subsection{SimREC}
  After the extensive ablations, we select the most effective designs for each REC component to form our strong base model, denoted as \emph{SimREC}.  The   detailed configurations are listed in Tab.~\ref{ablations}.

\begin{table*}[t]
	\centering
	\caption{ Ablation studies of different factors on three REC datasets. $\dagger$ denotes the settings of the basic structure and $\ddagger$ denotes the settings of SimREC. ``DN53'', ``CDN53'', ``Swin-B'' and ``ViT-B'' refer to the visual backbone of DarkNet-53~\cite{redmon2018yolov3:},  CSPDarkNet-53~\cite{ge2021yolox}, Swin-Transformer (Base)~\cite{liu2021swin} and Vision Transformer (Base)~\cite{dosovitskiy2020image}, respectively.  For all the results,
		we average three  experimental results of different random seeds.}
	%	\vspace*{-3mm} 
		\setlength\tabcolsep{7pt}
	\begin{tabular}{c|l|cccccc|cccc}
		\toprule
		\multirow{3}{*}{Factors}                                                         & \multicolumn{1}{c|}{\multirow{3}{*}{Choices}} & \multicolumn{2}{c}{RefCOCO} & \multicolumn{2}{c}{RefCOCO+} & \multicolumn{2}{c|}{RefCOCOg} & \multicolumn{2}{c}{\multirow{2}{*}{\begin{tabular}[c]{@{}c@{}}Inference\\ Speed{\footnotemark[4]}\end{tabular}}} & \multicolumn{2}{c}{\multirow{2}{*}{\begin{tabular}[c]{@{}c@{}}Training\\ Time\end{tabular}}} \\
		& \multicolumn{1}{c|}{}                         & \multicolumn{2}{c}{val}     & \multicolumn{2}{c}{val}      & \multicolumn{2}{c|}{val}      & \multicolumn{2}{c}{}                                                                           & \multicolumn{2}{c}{}                                                                         \\ \cline{3-12} 
		& \multicolumn{1}{c|}{}                         & Acc         & $\Delta$      & Acc         & $\Delta$       & Acc          & $\Delta$       & FPS                                          & $\Delta$                                        & Hours                                        & $\Delta$                                      \\ \hline
		\multirow{4}{*}{\begin{tabular}[c]{@{}c@{}}Visual\\ Backbone\end{tabular}}       & DN53$\dagger$                           & 70.63       & +0.00         & 50.39       & +0.00          & 58.78        & +0.00          & 59.2                                         & +0.0                                            & 6.6 h                                        & +0.0h                                         \\
		& VGG16                                        & 70.12       & -0.51         & 49.81       & -0.58          & 58.22        & -0.56          & 46.9                                         & -12.3                                           & 9.9 h                                        & +3.3h   
  \\
		& CDN53$\ddagger$                       & 71.90       & +1.17         & 57.24       & +6.85          & 57.58        & -1.20          & 74.1                                         & +14.9                                           & 5.4 h                                        & -1.2h    \\
  &ViT-B&72.60 &1.97 &58.12&+7.73&60.19&+1.41&30.6&-28.6& 10.1h &+3.5h\\
  & Swin-B&75.10&+4.47 &	59.96 & +9.57	&62.03&+3.25&36.2&-23.0& 9.7h&+3.1h \\ \hline
		\multirow{6}{*}{\begin{tabular}[c]{@{}c@{}}Language\\ Encoder\end{tabular}}      & LSTM+GLOVE$\dagger$                           & 70.63       & +0.00         & 50.39       & +0.00          & 58.78        & +0.00          & 59.2                                         & +0.0                                            & 6.6 h                                        & +0.0h                                         \\
		& LSTM                                          & 70.06       & -0.57         & 49.54       & -0.85          & 57.78        & -1.00          & 59.2                                         & +0.0                                            & 6.6 h                                        & +0.0h                                         \\
		& LSTM+GLOVE+SA(1)$\ddagger$                    & 71.56       & +0.93         & 51.26       & +0.87          & 58.11        & -0.67          & 58.1                                         & -1.1                                            & 6.7 h                                        & +0.1h                                         \\
		& LSTM+GLOVE+SA(2)                              & 71.17       & +0.54         & 50.65       & +0.26          & 58.42        & -0.36          & 57.1                                         & -2.1                                            & 6.7 h                                        & +0.1h                                         \\
		& LSTM+GLOVE+SA(3)                              & 71.43       & +0.80         & 48.73       & -1.66          & 59.15        & +0.37          & 55.9                                         & -3.3                                            & 6.7 h                                        & +0.1h                                         \\
		& LSTM+BERT                                     & 70.66       & +0.03         & 50.27       & -0.12          & 60.31        & +1.53          & 42.0                                         & -17.2                                           & 7.8 h                                        & +1.2h          
   \\
		& LSTM+RoBERTa                                     & 70.84       & +0.21         & 50.44       & -0.05          & 60.56        & +1.78          & 41.3                                         & -17.9                                           & 7.9 h                                        & +1.3h   \\
  		& LSTM+ALBERT                                     & 71.44       & +0.81         & 50.76       & +0.37         & 60.44        & +1.66          & 51.1                                         & -8.1                                           & 7.4 h                                        & +0.6h   
  \\ \hline
		\multirow{4}{*}{\begin{tabular}[c]{@{}c@{}}Multi-scale\\ Features\end{tabular}}  & One-scale Feature$\dagger$                    & 70.63       & +0.00         & 50.39       & +0.00          & 58.78        & +0.00          & 59.2                                         & +0.0                                            & 6.6 h                                        & +0.0h                                         \\
		& Two-scale Features                            & 76.79       & +6.16         & 63.71       & +13.32         & 62.62        & +3.84          & 54.1                                         & -5.1                                            & 7.9 h                                        & +1.3h                                         \\
		& Three-scale Features$\ddagger$                & 77.41       & +6.78         & 64.72       & +14.33         & 65.51        & +6.73          & 49.0                                         & -10.2                                           & 10.8 h                                       & +4.2h                                         \\
		& Four-scale Features                           & 78.33       & +7.70         & 65.21       & +14.82         & 65.56        & +6.78          & 42.9                                         & -16.3                                           & 11.1 h                                       & +4.5h                                         \\ \hline
		\multirow{3}{*}{\begin{tabular}[c]{@{}c@{}}Multi-scale\\ Detection\end{tabular}} & One-scale Feature+Head$\times$1$\dagger$      & 70.63       & +0.00         & 50.39       & +0.00          & 58.78        & +0.00          & 59.2                                         & +0.0                                            & 6.6 h                                        & +0.0h                                         \\
		& Three-scale Features+Head$\times$1$\ddagger$   & 77.41       & +6.78         & 64.72       & +14.33         & 65.51        & +6.73          & 49.0                                         & -10.2                                           & 10.8 h                                       & +4.2h                                         \\
		& Three-scale Features+Head$\times$3             & 77.40       & +6.77         & 62.78       & +12.39         & 64.73        & +5.95          & 39.2                                         & -20.0                                           & 14.5 h                                       & +7.9h                                         \\ \hline
		\multirow{2}{*}{\begin{tabular}[c]{@{}c@{}}Detection\\ Head\end{tabular}}        & Anchor-Based$\dagger$                         & 70.63       & +0.00         & 50.39       & +0.00          & 58.78        & +0.00          & 59.2                                         & +0.0                                            & 6.6 h                                        & +0.0h                                         \\
		& Anchor-Free$\ddagger$                         & 73.65       & +3.02         & 53.49       & +3.10          & 59.46        & +0.68          & 55.9                                         & -3.3                                            & 9.1 h                                        & +2.5h                                         \\ \hline
		\multirow{6}{*}{\begin{tabular}[c]{@{}c@{}}Loss\\ Fuction\end{tabular}}          & BCE+MSE$\dagger$                              & 70.63       & +0.00         & 50.39       & +0.00          & 58.78        & +0.00          & 59.2                                         & +0.0                                            & 6.6 h                                        & +0.0h                                         \\
		& Focal loss+MSE                                & 70.43       & -0.20         & 49.86       & -0.53          & 58.42        & -0.36          & 59.2                                         & +0.0                                            & 6.7 h                                        & +0.1h                                         \\
		& BCE+SmoothL1                                  & 71.12       & +0.49         & 50.29       & -0.10          & 59.23        & +0.45          & 59.2                                         & +0.0                                            & 6.7 h                                        & +0.1h                                         \\
		& BCE+IoU loss$\ddagger$                        & 71.39       & +0.76         & 51.68       & +1.29          & 59.54        & +0.76          & 59.2                                         & +0.0                                            & 6.7 h                                        & +0.1h                                         \\
		& BCE+GIoU loss                                 & 70.76       & +0.13         & 50.34       & -0.05          & 59.42        & +0.64          & 59.2                                         & +0.0                                            & 6.7 h                                        & +0.1h                                         \\ 
  & Reward loss\footnotemark[3] & 69.77 &-0.86& 50.89&+0.60& 60.10 & +1.32& 15.6 &-43.6 & 17.2h &+10.5 \\  \hline
		\multirow{5}{*}{\begin{tabular}[c]{@{}c@{}}Input\\ \\ Resolution\end{tabular}}   & $416 \times 416$ $\dagger$ $\ddagger$         & 70.63       & +0.00         & 50.39       & +0.00          & 58.78        & +0.00          & 59.2                                         & +0.0                                            & 6.6 h                                        & +0.0h                                         \\
		& $256 \times 256$                              & 64.57       & -6.06         & 47.25       & -3.14          & 52.63        & -6.15          & 69.0                                         & +9.8                                            & 3.6 h                                        & -3.0h                                         \\
		& $320 \times 320$                              & 68.53       & -2.10         & 49.13       & -1.26          & 56.86        & -1.92          & 67.1                                         & +7.9                                            & 5.0 h                                        & -1.6h                                         \\
		& $512 \times 512$                              & 70.97       & +0.34         & 51.28       & +0.89          & 59.97        & +1.19          & 46.9                                         & -12.3                                           & 9.6 h                                        & +3.0h                                         \\
		& $608 \times 608$                              & 71.15       & +0.52         & 51.59       & +1.20          & 59.38        & +0.60          & 41.0                                         & -18.2                                           & 12.6 h                                       & +6.0h                                         \\ \hline
		\multirow{10}{*}{\begin{tabular}[c]{@{}c@{}}Data\\ Augmentations\end{tabular}}   & None$\dagger$                                 & 70.63       & +0.00         & 50.39       & +0.00          & 58.78        & +0.00          & 59.2                                         & +0.0                                            & 6.6 h                                        & +0.0h                                         \\
		& Horizontal Flip                               & 66.58       & -4.05         & 47.92       & -2.47          & 56.13        & -2.65          & 59.2                                         & +0.0                                            & 6.6 h                                        & +0.0h                                         \\
		& Random Crop                                   & 61.20       & -9.43         & 47.06       & -3.33          & 51.67        & -7.11          & 59.2                                         & +0.0                                            & 6.8 h                                        & +0.2h                                         \\
		& Elastic Transform                             & 71.91       & +1.28         & 52.36       & +1.97          & 60.58        & +1.80          & 59.2                                         & +0.0                                            & 6.7 h                                        & +0.1h                                         \\
		& Random Resize$\ddagger$                       & 74.07       & +3.44         & 55.61       & +5.22          & 62.79        & +4.01          & 59.2                                         & +0.0                                            & 7.8 h                                        & +1.2h                                         \\
		& RandAugment                                   & 72.89       & +2.26         & 53.23       & +2.84          & 61.85        & +3.07          & 59.2                                         & +0.0                                            & 7.6 h                                        & +1.0h                                         \\
		& Random Erasing                                & 71.29       & +0.66         & 51.79       & +1.40          & 59.62        & +0.84          & 59.2                                         & +0.0                                            & 7.4 h                                        & +0.8h                                         \\
		& Mixup                                         & 72.63       & +2.00         & 51.09       & +0.70          & 59.56        & +0.78          & 59.2                                         & +0.0                                            & 7.2 h                                        & +0.6h                                         \\
		& CutMix                                        & 72.42       & +1.79         & 52.40       & +2.01          & 60.42        & +1.64          & 59.2                                         & +0.0                                            & 7.2 h                                        & +0.6h                                         \\
		& VG Data$\ddagger$                                 & 78.23       & +7.60          & 65.18       & +14.79         & 70.40        & +11.62         & 59.2                                         & +0.0                                            & 118.3h                                       & +111.7h                                       \\ \hline
		\multicolumn{1}{l|}{\multirow{4}{*}{Optimizations}}                              & None$\dagger$                                 & 70.63       & +0.00         & 50.39       & +0.00          & 58.78        & +0.00          & 59.2                                         & +0.0                                            & 6.6 h                                        & +0.0h                                         \\
		\multicolumn{1}{l|}{}                                                            & Label Smoothing                               & 70.78       & +0.15         & 50.93       & +0.54          & 59.27        & +0.49          & 59.2                                         & +0.0                                            & 6.6 h                                        & +0.0h                                         \\
		\multicolumn{1}{l|}{}                                                            & Exponential Moving Average$\ddagger$          & 71.95       & +1.32         & 51.78       & +1.39          & 60.46        & +1.68          & 59.2                                         & +0.0                                            & 5.1 h                                        & -1.5h                                         \\
		\multicolumn{1}{l|}{}                                                            & Learning Rate Cosine Decay$\ddagger$          & 70.79       & +0.16         & 50.63       & +0.30          & 59.44        & +0.66          & 59.2                                         & +0.0                                            & 6.6 h                                        & +0.0h                                         \\ \bottomrule
	\end{tabular}
	\label{ablations}
		\vspace{-4mm}
\end{table*}

\begin{table}[t]
	\centering
	\caption{Results of  combining different data augmentations.}
	%	\vspace{-2mm} 
		\setlength\tabcolsep{15pt}
	\begin{tabular}{c|cc}
		\toprule
		\multirow{2}{*}{\begin{tabular}[c]{@{}c@{}}Data\\ Augmentations\end{tabular}} & \multicolumn{2}{c}{RefCOCO (val)} \\ \cline{2-3} 
		& Acc            & $\Delta$         \\ \hline
		Random Resize (Baseline)&74.07&+0.00\\ \hline
		{MixUp}+CutMix                                                                  & 72.61          & -1.46            \\
		{RandAugment}+Elastic Transform                                                 & 71.99          & -2.08             \\
		RandAugment+{Random Resize       }                                              & 73.77          & -1.00            \\
		MixUp+{Random Resize  }                                                         & 71.51          & -2.56            \\ \hline
		All Useful Augmentations                                                    & 66.08          & -7.99            \\ \bottomrule
	\end{tabular}
		\vspace{-1.5em}
	\label{aug}
\end{table}

\begin{table*}[t]
	\centering
	\caption{Cumulative Ablations of SimREC on three REC datasets. The basic structure adopts the default setting of TABLE I.  Each row represents the performance after applying the setting of SimREC. }
	%	\vspace*{-3mm}
	\setlength\tabcolsep{3pt}
	\begin{tabular}{l|c|lll|lll|ll|l}
		\toprule
		\multicolumn{1}{c|}{\multirow{2}{*}{Models}} & Net      & \multicolumn{3}{c|}{RefCOCO}                                                     & \multicolumn{3}{c|}{RefCOCO+}                                                    & \multicolumn{2}{c|}{RefCOCOg}                       & \multicolumn{1}{c}{Inference} \\
		\multicolumn{1}{c|}{}                        & \#Params{\footnotemark[5]} & \multicolumn{1}{c}{val} & \multicolumn{1}{c}{testA} & \multicolumn{1}{c|}{testB} & \multicolumn{1}{c}{val} & \multicolumn{1}{c}{testA} & \multicolumn{1}{c|}{testB} & \multicolumn{1}{c}{val} & \multicolumn{1}{c|}{test} & \multicolumn{1}{c}{Speed{\footnotemark[4]}}     \\ \hline
		Basic Structure                              & 7M       & 70.63 \mypm{+0.00}      & 72.28 \mypm{+0.00}        & 66.50 \mypm{+0.00}         & 50.39 \mypm{+0.00}      & 52.93 \mypm{+0.00}        & 45.12 \mypm{+0.00}         & 58.78 \mypm{+0.00}      & 58.58 \mypm{+0.00}        & 58.8 fps \mypm{+0.0 fps}      \\
		+ Three-scale Features                       & 12M      & 77.41 \mypm{+6.78}      & 81.14 \mypm{+8.86}        & 71.85 \mypm{+5.35}         & 64.72 \mypm{+14.33}     & 69.09 \mypm{+16.16}       & 59.49 \mypm{+14.37}        & 65.51 \mypm{+6.73}      & 64.82 \mypm{+6.24}        & 50.0 fps \mypm{-8.8 fps}      \\
		+ LSTM-GLOVE-SA(1)                           & 16M      & 78.21 \mypm{+7.58}      & 81.22 \mypm{+8.94}        & 71.94 \mypm{+5.44}         & 64.83 \mypm{+14.44}     & 70.22 \mypm{+17.29}       & 55.23 \mypm{+10.11}        & 66.65 \mypm{+7.87}      & 66.57 \mypm{+7.99}        & 48.8 fps \mypm{-10.0 fps}     \\
		+ Anchor-free Head                           & 16M      & 79.74 \mypm{+9.11}      & 82.67 \mypm{+10.39}       & 73.56 \mypm{+7.06}         & 68.25 \mypm{+17.86}     & 73.12 \mypm{+20.19}       & 58.45 \mypm{+13.33}        & 69.22 \mypm{+10.44}     & 69.18 \mypm{+10.60}       & 47.9 fps \mypm{-10.9 fps}     \\
		+  Optimizations                      & 16M      & 80.88 \mypm{+10.25}     & 83.23 \mypm{+10.95}       & 75.22 \mypm{+8.72}         & 68.57 \mypm{+18.18}     & 73.65 \mypm{+20.72}       & 58.79 \mypm{+13.67}        & 69.26 \mypm{+10.48}     & 69.45 \mypm{+10.87}       & 47.9 fps \mypm{-10.9 fps}     \\
		+ Random Resize                              & 16M      & 82.03 \mypm{+11.40}     & 85.01 \mypm{+12.73}       & 77.65 \mypm{+11.15}        & 70.24 \mypm{+19.85}     & 75.87 \mypm{+22.94}       & 60.87 \mypm{+15.75}        & 71.43 \mypm{+12.65}     & 71.21 \mypm{+12.63}       & 47.9 fps \mypm{-10.9 fps}     \\
		+ CSPDarknet                                 & 16M      & 82.45 \mypm{+11.82}     & 85.91 \mypm{+13.63}       & 77.98 \mypm{+11.48}        & 70.58 \mypm{+20.19}     & 76.75 \mypm{+23.82}       & 61.12 \mypm{+16.00}        & 72.59 \mypm{+13.81}     & 72.86 \mypm{+14.28}       & 55.6 fps \mypm{-3.2 fps}      \\
		+ VG Data                                    & 16M      & 86.49 \mypm{+15.86}     & 88.95 \mypm{+16.67}       & 81.74 \mypm{+15.24}        & 77.51 \mypm{+27.12}     & 82.68 \mypm{+29.75}        & 68.81 \mypm{+23.69}        & 79.88 \mypm{+21.10}     & 79.56 \mypm{+20.98}       & 55.6 fps \mypm{-3.2 fps}      \\ \bottomrule
	\end{tabular}
	\label{cablations}
	%	\vspace{-3mm}
\end{table*}

\begin{table*}[t]
	\caption{Comparison of SimREC and the state-of-the-arts on three REC datasets.  SimREC (scratch) is not pre-trained on VG. TransVG+SimREC denotes that the findings of SimREC are applied to TransVG. {$\ddagger$ denotes that model is fine-tuned on the combination of three REC datasets.     The result colored in gray denotes that the model is trained with multi-task supervised learning, which requires more labeled data. The numbers in bold denote the best performance, while the underlined numbers are the ones among the compared methods.   }  {All inference speeds are tested on an NVIDIA 1080Ti (11GB). }  }
		\vspace*{-3mm} 
		\setlength\tabcolsep{6pt}
	\begin{center}
		{%
			\begin{tabular}{l|c cc| c c c | c c c | c c|c}
				\toprule
				\multirow{2}{*}{Models}  & Visual &Pretrain  & Net & \multicolumn{3}{c|}{RefCOCO} & \multicolumn{3}{c|}{RefCOCO+} & \multicolumn{2}{c|}{RefCOCOg}& Inference \\ 
				& Features & Images & \#Params{\footnotemark[5]} & val & testA & testB & val & testA & testB & val-u & test-u & \multicolumn{1}{c}{Speed{\footnotemark[4]}} \\  \hline
				\multicolumn{1}{l|}{\textit{Two-stage:}} & & & & & & & & && \\
				CMN~\cite{hu2017modeling}\mypm{CVPR17}  & VGG16  & -& - & - & 71.03 & 65.77 & - & 54.32 & 47.76 & - & -&- \\
				MAttNet~\cite{MATT:}\mypm{CVPR18} &	RN101& -  & 18M & 76.65 & 81.14 & 69.99 & 65.33 & 71.62 & 56.02 & 66.58 & 67.27 &2.6 fps\\
				%RvG-Tree~\cite{}  & RN101 & None & $$ & 75.06 & 78.61 & 69.85 & 63.51 & 67.45 & 56.66 & 66.95 & 66.51 \\
				NMTree~\cite{liu2019learning}\mypm{ICCV19} & RN101& -  & - & 76.41 & 81.21 & 70.09 & 66.46 & 72.02 & 57.52 & 65.87 & 66.44&- \\
				CM-Att-Erase~\cite{liu2019improving}\mypm{CVPR19} &	RN101& - & - & 78.35 & 83.14 & 71.32 & 68.09 & 73.65 & 58.03 & 67.99 & 68.67&-\\ \hline
				\multicolumn{1}{l|}{\textit{One-stage:}} & & & & & & & & & && \\
				RealGIN~\cite{zhou2021real}\mypm{TNNLS21} &DN53& -&12M& 77.25 &78.70& 72.10& 62.78& 67.17& 54.21&62.75& 62.33&\underline{52.6  fps}\\
				%RCCF~\cite{} & DLA34 & None & $$ & - & 81.06 & 71.85 & - & 70.35 & 56.32 & - & 65.73 \\
				%SSG~\cite{}  & DN53 & None & $$ & - & 76.51 & 67.50 & - & 62.14 & 49.27 & 58.80 & - \\  
				FAOA~\cite{yang2019fast}\mypm{ICCV19} & DN53 & - & 120M & 72.54 & 74.35 & 68.50 & 56.81 & 60.23 & 49.60 & 61.33 & 60.36&25.0 fps \\
				ReSC~\cite{ReSC}\mypm{ECCV20} & DN53& -  & 120M & 77.63 & 80.45 & 72.30 & 63.59 & 68.36 & 56.81 & 67.30 & 67.20 & 18.9 fps\\
				MCN~\cite{MCN}\mypm{CVPR20}& DN53 & - &26M & 80.08 & 82.29 & 74.98 & 67.16 & 72.86 & 57.31 & 66.46 & 66.01& 20.3 fps \\
				Iter-Shrinking~\cite{sun2021iterative}\mypm{CVPR21} &RN101& -&-&-&74.27&68.10&-&71.05&58.25&-&{70.05}&-\\
				 PFCOS~\cite{sun2022proposal}\mypm{TMM22} &DN53& -&- &79.50& 81.49& 77.13& 65.76& 69.61& 60.30 &{69.06}& 68.34&-\\
				LBYL-Net~\cite{huang2021look}\mypm{CVPR21}&DN53& -&115M&79.67 &82.91 &74.15& {68.64} &{73.38}& {59.49}  &-&-&27.0  fps\\
				TransVG~\cite{deng2021transvg}\mypm{ICCV21} &RN101& -&117M&{81.02} &{82.72}& {78.35}& 64.82 &70.70 &56.94& {68.67}& 67.73&19.6 fps\\
    				TransVG~\cite{deng2021transvg}+SimREC  &CDN53& -&117M&{82.23} &{85.24}& {79.85}& 68.52 & {73.42} & {60.11}& {70.03}& 69.88&27.5 fps\\
        One-for-All~\cite{zhang2023one}\mypm{Neurocomputing23} &DN53&-&-&76.99 &79.71 &72.67 &61.58 &66.60 &54.00 &66.03 &66.70&-\\
                SeqTR~\cite{zhu2022seqtr} \mypm{ECCV22} &DN53&-& 19M&81.23& 85.00 &76.08 &68.82& 75.37 &58.78 &71.35 &71.58 &21.2 fps\\
                PLV-FPN~\cite{liao2022progressive}\mypm{TIP22} &RN101& -&-& 81.93 &84.99& 76.25& 71.20 &77.40& 61.08 &70.45& 71.08&-\\
                Word2Pix~\cite{zhao2022word2pix}\mypm{TNNLS22}&RN101& -&-& 81.20& 84.39 &78.12 &69.74& 76.11 &61.24 &70.81 &71.34&-\\
                 QRNet~\cite{ye2022shifting}\mypm{CVPR22} & Swin-S& -& 117.9M &\underline{84.01} &\underline{85.85} &\underline{82.34}& \underline{72.94} &\underline{{76.17}} &\underline{63.81} &\underline{71.89} &\underline{73.03}&20.1 fps\\
			\rowcolor{lightgray}			SimREC (ours)  & DN53 & - & 16M&82.03&85.01&77.65&70.24&75.39&61.85&71.43&72.07&50.0 fps\\ %ckpt on cedar
				\rowcolor{lightgray}		SimREC  (ours)   & CDN53 & - & 16M&82.45&85.91&77.98&	70.58&\textbf{76.75}&61.12	&72.59&72.86
				&\textbf{55.6 fps}\\ % ckpt on edith
    				\rowcolor{lightgray}		SimREC  (ours) & Swin-B & - & 16M&\textbf{84.93} &	\textbf{86.55}	&\textbf{83.85}	&\textbf{73.09}&	{76.02}	&\textbf{66.28}	&\textbf{75.90}	&\textbf{75.55}
				& 36.2 fps\\ \hline % ckpt on edith\ % ckpt on edith
	\multicolumn{1}{l|}{\textit{Additional REC data:}} & & & & & & & & & && \\MDETR\cite{kamath2021mdetr}\mypm{ICCV21}&RN101&0.2M&143M&\underline{86.75}& \underline{89.58}& \underline{81.41}& \underline{79.52}& \underline{84.09}& \underline{70.62}& \underline{81.64}& \underline{80.89}&\underline{15.4 fps}\\	UniTAB~\cite{yang2022unitab}\mypm{ECCV22}&RN101&0.2M&-&	86.32& 88.84 &80.61& 78.70 &83.22& 69.48& 79.96 &79.97&-\\
     \color{gray} SeqTR~\cite{zhu2022seqtr}\mypm{ECCV22}&  \color{gray} DN53&\color{gray} 0.2M&  \color{gray} 19M&    \color{gray} 87.00&  \color{gray}  90.15&  \color{gray} 83.59&  \color{gray} 78.69&  \color{gray} 84.51&  \color{gray} 71.87&  \color{gray} 82.69&  \color{gray} 83.37&  \color{gray} 21.2 fps\\
  \color{gray}  UNINEXT$\ddagger$~\cite{yan2023universal}\mypm{CVPR23}&  \color{gray} RN50&  \color{gray} 0.6M&-&  \color{gray}  89.72&   \color{gray} 91.52&   \color{gray} 86.93&   \color{gray} 79.76&  \color{gray}  85.23&   \color{gray} 72.78&   \color{gray} 83.95&   \color{gray} 84.31 &  \color{gray} -\\
    \color{gray}   PolyFormer-L$\ddagger$~\cite{liu2023polyformer}\mypm{CVPR23} &   \color{gray} Swin-L&  \color{gray}  0.2M&  \color{gray} 363M&   \color{gray} 90.38 &  \color{gray} 92.89 &   \color{gray} 87.16 &  \color{gray}  84.98&   \color{gray} 89.77 &  \color{gray} 77.97&   \color{gray} 85.83 &  \color{gray} 85.91&   \color{gray} 2.5 fps\\
%       \color{gray} OFA-L$\ddagger$~\cite{wang2022ofa}\mypm{ICML22}&  \color{gray} RN152&  \color{gray} 60.6M&\color{gray} 412M&  \color{gray} 90.05&  \color{gray}  92.93&   \color{gray} 85.26&   \color{gray} 85.80 &  \color{gray} 89.87 &  \color{gray} 79.22&   \color{gray} 85.89&   \color{gray} 86.55 & \color{gray} 
% 2.7 fps
% \\
          		\rowcolor{lightgray}	           SimREC&Swin-B&0.2M &16M&88.80	&89.48&	\textbf{87.58}	&78.94&	82.69	&{72.80}	&82.84&	83.41& 30.6 fps\\
			\rowcolor{lightgray}		SimREC &CDN53&0.2M &16M&{88.47}&{90.75}&{84.10}&{80.25}&{85.33}&{72.04}&	{83.08}&{83.99}&{55.6 fps}\\
  	\rowcolor{lightgray}  SimREC$\ddagger$&CDN53&0.2M &16M&\textbf{89.39}&	\textbf{92.12	}&86.06&	\textbf{81.42}	&\textbf{86.74}&	\textbf{74.13}&	\textbf{83.64}	&\textbf{85.02} &\textbf{55.6 fps}\\   
				\bottomrule  
		\end{tabular}}
	\end{center}
	\label{tab:refcoco_results}
	%	\vspace{-3em}
\end{table*}

\begin{table*}[t]
	%	\caption{{\bf Comparison on REC task.} Performance on  RefCOCO/RefCOCO+/RefCOCOg datasets \cite{yu2016modeling} is reported. Ours$^*$ denotes that pretraining is used. RN50 and RN101 refer to ResNet50 and ResNet101 \cite{he2016deep} respectively; DN53 refers to DarkNet53 \cite{redmon2018yolov3} backbone.}
	\centering
	\caption{Comparison of SimREC and large-scale BERT-style pre-trained models on three REC datasets. $\ddagger$ denotes that model is fine-tuned on the combination of three REC datasets.   The result colored in gray denotes that the model uses much more labeled data. }
		\vspace*{-3mm}
	\setlength\tabcolsep{6.5pt} 
	\begin{center}
		{%
			\begin{tabular}{c| c|c|c| c c c | c c c | c c|c}
				\toprule
				\multirow{2}{*}{Models}  & Visual & Pretrain & Net & \multicolumn{3}{c|}{RefCOCO} & \multicolumn{3}{c|}{RefCOCO+} & \multicolumn{2}{c|}{RefCOCOg}& Inference \\ 
				& Features & Images & \#Params{\footnotemark[5]} & val & testA & testB & val & testA & testB & val-u & test-u &\multicolumn{1}{c}{Speed{\footnotemark[4]}} \\
				\hline   
				VilBERT\cite{lu2019vilbert}\mypm{NeurIPS19} & RN101 & 3.3M & 221M & - & - & - & 72.34 & 78.52 & 62.61 & - & - &{2.5 fps}\\
				ERNIE-ViL-L\cite{yu2020ernie}\mypm{AAAI21} & RN101 & 4.3M & -& - & - & - & 75.89 & {82.37} &  \underline{66.91} & - & - &-\\
				UNITER-L\cite{chen2019uniter}\mypm{ECCV20} & RN101 & 4.6M & 335M & 81.41 & 87.04 & 74.17 & 75.90 & 81.45 & 66.70 & 74.86 & 75.77 &2.4 fps\\
				VILLA-L\cite{gan2020large}\mypm{NeurIPS20} & RN101 & 4.6M & 335M  &  \underline{82.39} & \underline{87.48} & \underline{74.84} & \underline{76.17} & \underline{81.54} & 66.84 & \underline{76.18} & \underline{76.71} &2.4 fps  \\ 
          \color{gray} OFA-B$\ddagger$~\cite{wang2022ofa}\mypm{ICML22}&  \color{gray} RN101&  \color{gray} 60.6M&\color{gray} 137M&  \color{gray} 88.48&  \color{gray}  90.67&   \color{gray} 83.30&   \color{gray} 81.39 &  \color{gray} 87.15 &  \color{gray} 74.29&   \color{gray} 82.29&   \color{gray} 82.31 & \color{gray} 
9.7 fps \\
      \color{gray} OFA-L$\ddagger$~\cite{wang2022ofa}\mypm{ICML22}&  \color{gray} RN152&  \color{gray} 60.6M&\color{gray} 412M&  \color{gray} 90.05&  \color{gray}  92.93&   \color{gray} 85.26&   \color{gray} 85.80 &  \color{gray} 89.87 &  \color{gray} 79.22&   \color{gray} 85.89&   \color{gray} 86.55 & \color{gray} 
2.7 fps \\ \hline 
				SimREC & DN53 & 0.1M &   16M &85.72&88.10&81.28&76.11&81.57&66.49&77.78&78.28&50.0 fps\\
				SimREC & CDN53 & 0.1M &  16M&{86.49}&{88.95}&{81.74}&{77.51}&{82.68}&{68.81}&{79.88}&{79.56}&{55.6 fps} \\
        SimREC&Swin-B&0.2M &16M&88.80	&89.48&	\textbf{87.58}	&78.94&	82.69	&{72.80}	&82.84&	83.41& 30.6 fps\\
				SimREC &CDN53&0.2M &16M&{88.47}&{90.75}&{84.10}&{80.25}&{85.33}&{72.04}&	{83.08}&{83.99}&{55.6 fps}\\
    SimREC$\ddagger$&CDN53&0.2M &16M&\textbf{89.39}&	\textbf{92.12	}&86.06&	\textbf{81.42}	&\textbf{86.74}&	\textbf{74.13}&	\textbf{83.64}	&\textbf{85.02} &\textbf{55.6 fps}\\ \bottomrule
		\end{tabular}}
	\end{center}
	\label{tab:pretrained_results}
	\vspace{-2em}
\end{table*}

\begin{table}[t]
	\centering 
	\caption{Comparison of SimREC and the state-of-the-arts on Referit and Flickr. }
	\setlength\tabcolsep{15pt}
	\begin{tabular}{l|ccc}
		\toprule
		\multicolumn{1}{c|}{\multirow{2}{*}{Models}} & \multirow{2}{*}{\begin{tabular}[c]{@{}c@{}}Visual\\ Features\end{tabular}} & Referit & Flickr \\
		\multicolumn{1}{c|}{}                        &                                                                            & test    & test  \\ \hline
		\textit{Two-stage:} & & & \\
		MAttNet~\cite{MATT:}                                     & RN101                                                                      & 29.04   & -     \\
		Similarity Net~\cite{wang2018learning}                              & RN101                                                                      & 34.54   & 60.89 \\
		DDPN~\cite{yu2018rethinking}                                        & RN101                                                                      & 63.00   & 73.30 \\ \hline
		\textit{One-stage:} & & & \\
		ZSGNet~\cite{sadhu2017zero}                                      & RN50                                                                       & 58.63   & 63.39 \\
		FAOA~\cite{yang2019fast}                                        & DN53                                                                       & 60.67   & 68.71 \\
		ReSC~\cite{ReSC}                                        & DN53                                                                       & 64.60   & 69.28 \\
		TransVG~\cite{deng2021transvg}                                     & RN101                                                                      & \underline{70.73}   & \underline{79.10}\\ \hline
		SimREC                                      & CDN53                                                                      & \textbf{77.80}   & \textbf{83.31} \\ \bottomrule
	\end{tabular}
	\vspace{-1em}
	\label{SOTA} 
\end{table}

\section{Experiments}
\subsection{Datasets and Metric}
\textbf{RefCOCO \& RefCOCO+}~\cite{REFCOCO}  contain  142k referring expressions for 50k bounding boxes in 19k images from MS-COCO~\cite{MSCOCO}. There are four splits in RefCOCO and RefCOCO+, \emph{i.,e,} \textit{train}, \textit{val}, \textit{testA} and \textit{testB}, with a number of 12k, 10k, 5k, 5k images, respectively. Test A contains more \emph{people} instances, while Test B has more \emph{object} ones.  The expressions of RefCOCO are mainly about absolute spatial relationships, while the ones of RefCOCO+ are about attributes and  relative relations. 
\textbf{RefCOCOg}~\cite{REFCOCOG,nagaraja2016modeling} has 104k expressions for 54k objects from 26k images.  It has two different partitions, \emph{i.e.}, \textit{Google} split~\cite{REFCOCOG} and {\textit{UMD}} split~\cite{nagaraja2016modeling}. Google split only contains  {training} set and {validation} set,  and  images of which are overlapped.   Instead, {\textit{UMD}} split  avoids  overlapping partitions,  and it contains three splits, \emph{i.e.,} \emph{train}, \emph{validation} and \emph{test}.  In particular, the expressions of RefCOCOg are longer and more complex than those of RefCOCO and RefCOCO+. \textbf{Visual Genome}~\cite{krishna2016visual} contains 5.6M referring expressions for 101K images. Compared to RefCOCO, RefCOCCO+ and RefCOCOg,   the expressions of Visual Genome are more diverse but their annotations are relatively noisy.

\textbf{Intersection-over-Union (IoU)} is the metric  used in REC, which   measures the overlap degree between  the prediction and the ground-truth.  Following previous works~\cite{MCN,zhou2021real,yang2019fast,ReSC}, we use \textit{IoU@0.5} to measure prediction accuracy.

\subsection{Experimental Settings}
\noindent\textbf{Network Configurations.} The visual backbone and the language encoder of SimREC are a CSPDarknet~\cite{wang2020cspnet} and an LSTM~\cite{LSTM} with a self-attention layer~\cite{vaswani2017attention}, respectively.  For SimREC, we use three-scale visual features from the  visual backbone outputs of stride=8, stride=16 and stride=32, of which dimensions are 256, 512, 1024, respectively. For language encoder, its dimension is set to 512 and the word embedding is a 300-d matrix initialized by pre-trained GLOVE vectors~\cite{pennington2014glove}.  
By default, the input image is resized to $416 \times 416$, and the maximum length of input expressions is set to 15 for RefCOCO and RefCOCO+, and 20 for RefCOCOg.  
For detection head and loss function, we apply the anchor-free head~\cite{ge2021yolox} with IoU loss~\cite{rezatofighi2019generalized} and BCE loss.  For data augmentations,  random resize is used in the training.  To improve the training efficiency, we further apply the exponential moving average (EMA)~\cite{tarvainen2017mean} and cosine decay learning schedule. %More network details such as dimensions can refer to our appendix.

\noindent\textbf{Training settings.}  We use Adam~\cite{ADAM} to train our models for 40 epochs, which   can be reduced to 25 epochs when EMA is applied.  The learning rate is initialized with 1$e$-4 and  will be decayed by 10 at the 35-th, 37-th and 39-th epoch, respectively. We also try the cosine decay as the learning rate schedule.   All visual backbones are pre-trained  on MS-COCO~\cite{MSCOCO},  while  the images appeared in the val and test sets of three REC datasets are removed. We also apply  the region annotations in the Visual Genome~\cite{VG} to pre-train SimREC.  Particularly, pre-trained stage takes 15 epochs with a batch size of 256.    The   learning rate and optimization are kept the default settings.

\subsection{Experimental Results}
\subsubsection{Ablation Study}
\label{ablation_text}
We  first conduct extensive experiments on benchmark datasets to ablate the candidate designs/settings described in Sec.~\ref{ablation_text}, of which results are shown in Tab.~\ref{ablations}. 

\noindent\textbf{What matters in one-stage REC?}  Form Tab.~\ref{ablations}, we  first observe that the visual components are critical for one-stage REC. The use of a better backbone or detection head can lead to significant performance gains. {In particular, Swin-Transformer~\cite{liu2021swin} provides the most significant gains, \emph{i.e.,} +9.6\%  on RefCOCO+. Nevertheless, such a large visual backbone also slows down the inference speed of the one-stage REC model.  In contrast, anchor-free head not only boosts the detection performance but also maintains  the inference efficiency. }  Meanwhile, larger image resolutions are also more beneficial for performance. These results are within the scope of existing CV and V\&L studies~\cite{jiang2020defense,luo2020cascade,ren2017faster,redmon2018yolov3:,zhang2019bag}.

Multi-scale visual features are the factor that can boost performance. On RefCOCO+, its performance gain can reach up to +14.3\%, while the ones on the other two datasets are also about +6\%. However, when combining it with multi-scale detection, the performance is not further improved. This result suggests that multi-scale features are basically used to enhance the descriptive power of visual backbones rather than multi-scale detection, which is obviously different to  the use in object detection~\cite{redmon2018yolov3:,ren2017faster}.

Data augmentation is another factor that greatly affects performance. Using Visual Genome   (VG) as the additional training data can greatly boost performance on all benchmarks. Meanwhile, a simple image augmentation like \emph{Resizing} also obtain obvious performance gains. These results suggest that one-stage REC has a great demand in training data, especially compared with other  V\&L tasks~\cite{VQA0,VQA1}.   For example, VG is often used as additional data in VQA, while the improvement is {often marginal}~\cite{MCAN,BUTD}.   
From Tab.~\ref{ablations}, we can also summarize a prerequisite for data augmentation,  which  is that the semantics of image-text pairs should not be damaged. 
The operations like \emph{Horizontal Flip and Random Crop} will reduce the completeness or the correctness of REC examples, thus leading to counterproductive results. Meanwhile, strong augmentation methods are inferior than the simple ones, \emph{e.g.}, RandomErasing and CutMix. When combing all positive augmentation methods, the REC performance encounters a significant decline, as shown in Tab.~\ref{aug},   which is opposite to that in object detection~\cite{zhang2019bag}. These results greatly differ REC from object detection in terms of data augmentation.

In Tab.~\ref{cablations}, we conduct  cumulative ablations to validate  the combination of SimREC settings.  From the results, we observe that each factor consistently improves  performance when combined with other factors. Particularly, multi-scale features and VG data augmentations still play the most critical role on performance, providing up to +12.94\% gains on RefCOCOg \textit{test} set. Meanwhile, the combination of other factors also brings notable performance gains. For example, language encoder, visual backbone, detection head, optimizations and random resize provide the improvements of +1.75\%, +1.65\%, +2.61\%, +0.27\% and +1.76\% on RefCOCOg \textit{test} set, respectively. And their combination finally improves the performance by +8.04\% on RefCOCOg \textit{test} set.  After combining all  factors, +20.98\% performance gains can be observed on RefCOCOg \textit{test} set.  These results further confirm the effectiveness of SimREC.

\footnotetext[3]{Reward loss is borrowed from Iter-Shrinking~\cite{sun2021iterative}, which  requires multiple forwards during inference.}
\noindent\textbf{What are less important in one-stage REC?} Compared with the above mentioned factors, the other components have less impact on one-stage REC. 
Above all, the role of language encoder is not as important as that in other V\&L tasks. 
By deploying the advanced designs like   BERT~\cite{devlin2018bert:} and ALBERT~\cite{lan2019albert}, the performance gains are not obvious. Even on RefCOCOg, of which expressions are long and complex, the improvement is only +1.78\%. One assumption about this result is that the expressions of REC is not so difficult to understand.  But more importantly, it may also suggest that REC is less affected by language priors, which {is often serious in other} V\&L tasks~\cite{VQA0,VQA1,kiros2014multimodal}.  In this case, a simple design can meet the requirement of expression comprehension. 

%这边Objective loss感觉写的不准确、专业，需要再CHECK一下
In Tab.~\ref{ablations}, we  evaluate the effects of different loss functions and training tricks. 
The loss functions  are deployed on the anchor-based detector~\cite{redmon2018yolov3:}, while their performance gains are not obvious,  especially compared with object detection~\cite{rezatofighi2019generalized}.  This results may suggest that   pre-defined anchors  are not  suitable for one-stage REC. { We also validate the effect of the reinforcement learning approach~\cite{sun2021iterative}, which  reduces the performance and the efficiency.}   In terms of model optimization, all candidate methods pose positive impacts on performance although not so obvious. Particularly, we find that EMA~\cite{tarvainen2017mean} can  accelerate training convergence.

\footnotetext[4]{Inference speed is tested on the 1080Ti (11GB).}

\begin{figure*}[t]
	\centering
	\includegraphics[width=2\columnwidth]{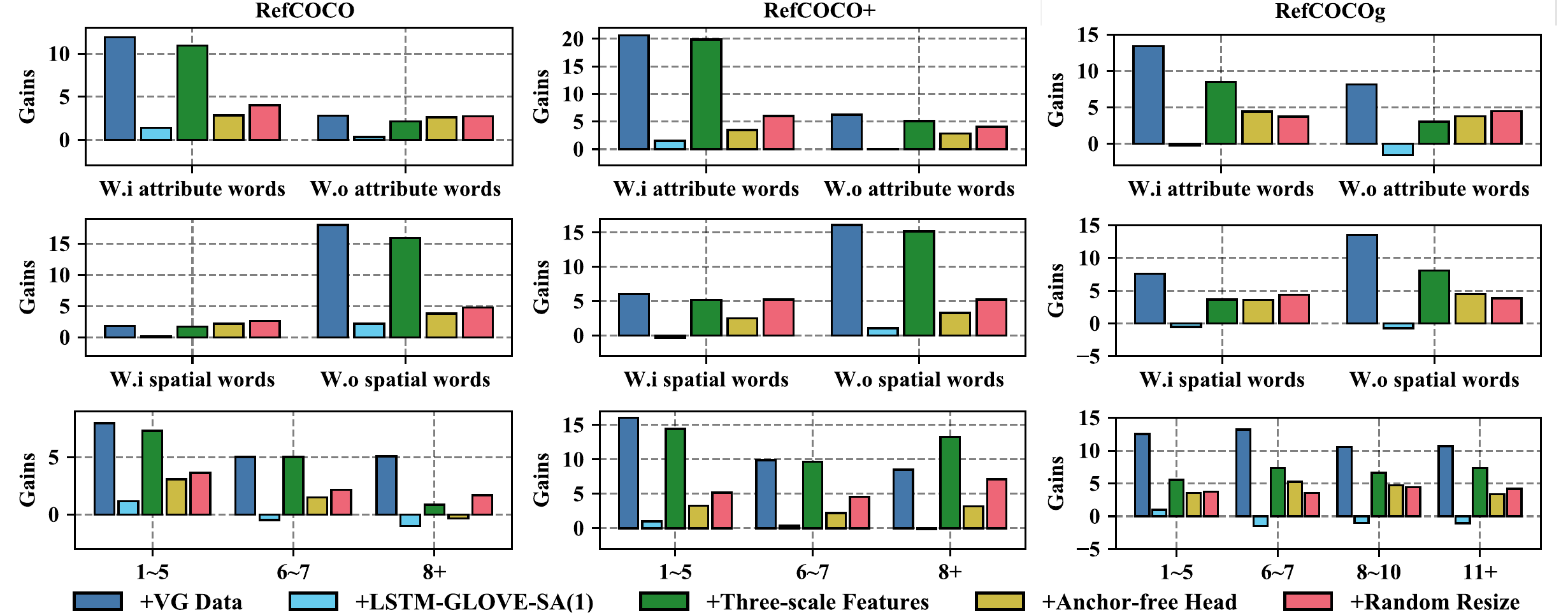}
	
	% Reduce the figure size so that it is slightly narrower than the column. Don't use precise values for figure width.This setup will avoid overfull boxes. 
	\caption{Relative performance gains of five settings   in SimREC  on  attribute descriptions  (row-1),  spatial descriptions (row-2), the expression length (row-3).   All results are calculated on the \textit{val} set.} 
	\label{fig2}
	%	\vspace{-2mm}
		\vspace{-1em}
\end{figure*}

\begin{figure*}[t]
	\centering
	\includegraphics[width=2\columnwidth]{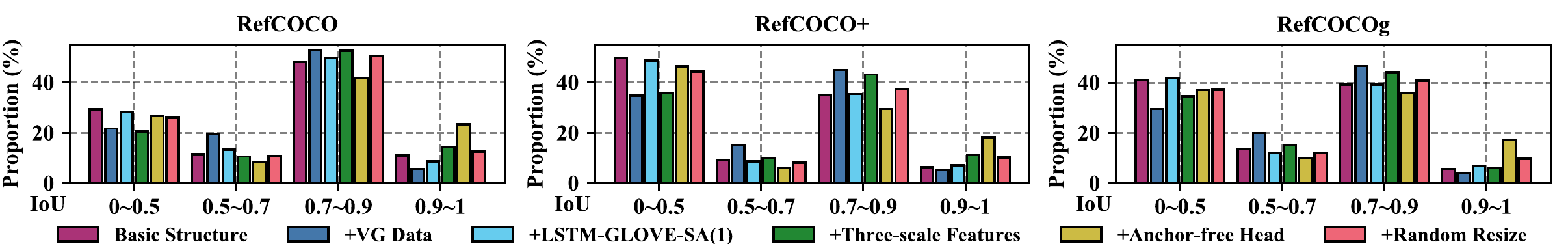}
	
	% Reduce the figure size so that it is slightly narrower than the column. Don't use precise values for figure width.This setup will avoid overfull boxes. 
	\caption{ Distribution of the predicted bounding boxes on different IoU  metrics. All results are calculated on the \textit{val} set.} 
	\label{fig3}
		\vspace{-1.5em}
	%	\vspace{-4mm}
\end{figure*}

\noindent\textbf{How do these factors affect REC?}  Based on the above findings, we further  explore how these key factors affect one-stage REC. Specifically, we investigate their effects on expressions of different content and lengths in Fig.~\ref{fig2}, and the quality of predicted bounding boxes  in Fig.~\ref{fig3}. 
%The studied factors include data augmentation, \emph{i.e.} Visual Genome~\cite{krishna2016visual} and \emph{Resizing}, multi-scale features, detection head (anchor-free head) and language encoder (LSTM+BERT). 

\footnotetext[5]{The calculation of network parameters excludes the visual backbone.}

From Fig.~\ref{fig2}, we can first observe that multi-scale features and data augmentations can improve the model performance on all expressions, especially the ones with attribute descriptions (words). In this regard, we think that multi-scale features are to enhance the visual representations to facilitate fine-grained recognition, while data augmentation can significantly strengthen the learning of language-vision alignments. These two aspects are critical for REC.
In addition, we also notice that the improvement of the expressions with spatial information (words) is  small. We attribute it to the  intrinsic merit  of one-stage REC modeling   in spatial relation modeling ~\cite{zhou2021real}. 
Meanwhile, we can also find that with the improvement in visual modeling, the model also has a better ability to process the long and complex expressions, especially with the help of VG. This result also provides a useful hint for the study of multi-modal reasoning. 

Compared with these two factors, the benefit of detection head is more balanced. For this end, we understand that detection head mainly affects the model's detection ability. Besides, Fig.~\ref{fig2} also reflects the inference of language encoder, which is relatively small. Since most of the sentences in benchmark datasets are short, the help of language encoder is not obvious. In contrast, it also affects its performance on long sentences. This observation is consistent with the  experimental results  in Tab.~\ref{ablations}. 

Fig.~\ref{fig3} depicts the IoU score distributions of the model's predictions about the ground-truth bounding boxes.  These results reflect the impact of these factors on the quality of model predictions. From this figure, we  observe that the anchor-free detector obviously improves the detection ability. On the high-quality detection (IoU 0.9-1), the number is almost doubled by this detector. Meanwhile, we  notice that the use of VG  helps the model improve the detection accuracy (IoU$>$0.5), but its performance on high-quality detection declines obviously. This is mainly attributed to the less  accurate bounding box  annotations in  VG~\cite{krishna2016visual}.

\begin{table}[t]
	\centering
	\caption{Results of applying our findings to Transformer-based REC model. ViLT is pre-trained on 3M image-text pairs. By default, we use the [cls] token to predict the box of the referent during fine-tuning. }
	\setlength\tabcolsep{15pt}
	\begin{tabular}{lccc}
		\toprule
		\multicolumn{4}{c}{RefCOCO}                              \\ \hline
		\multicolumn{1}{l|}{}            & val   & testA & testB \\ \hline
          	\multicolumn{1}{l|}{TransVG~\cite{deng2021transvg}}& 81.02&82.72&78.35\\
		\multicolumn{1}{l|}{ViLT~\cite{kim2021vilt}}        & 79.51 & 82.15 & 71.93 \\ 
          	\multicolumn{1}{l|}{TransVG+SimREC}& 82.52&84.12&\textbf{79.85}	\\
		\multicolumn{1}{l|}{ViLT+SimREC} &  \textbf{83.23 }    & \textbf{85.24}      & 79.12     \\\bottomrule \toprule
		\multicolumn{4}{c}{RefCOCO+}                             \\ \hline
		\multicolumn{1}{l|}{}            & val   & testA & testB \\ \hline
  \multicolumn{1}{l|}{TransVG~\cite{deng2021transvg}}& 64.82&70.70&56.94\\
		\multicolumn{1}{l|}{ViLT~\cite{kim2021vilt}}        & 65.32 & 71.30 & 54.63 \\
            	\multicolumn{1}{l|}{TransVG+SimREC}& 68.52&73.42& 60.11 \\
		\multicolumn{1}{l|}{ViLT+SimREC} &  \textbf{73.70 }    &  \textbf{78.54 }    &  \textbf{63.57}     \\ \bottomrule \toprule
		\multicolumn{4}{c}{RefCOCOg}                             \\ \hline
		\multicolumn{1}{l|}{}            & val   & test  &       \\ \hline
  \multicolumn{1}{l|}{TransVG~\cite{deng2021transvg}}& 68.67&67.73 \\
		\multicolumn{1}{l|}{ViLT~\cite{kim2021vilt}}        & 67.03 & 66.35 &       \\
    \multicolumn{1}{l|}{TransVG~\cite{deng2021transvg}+SimREC}& 70.03&69.88 \\
		\multicolumn{1}{l|}{ViLT+SimREC} &    \textbf{73.53}   &   \textbf{74.05}    &       \\ \bottomrule
	\end{tabular}
	\vspace{-2em}
\label{gen}
\end{table}


%In this paragraph, we seek to understand how the effective factors, \emph{i.e.,} visual backbone, language encoder, multi-scale features, detection head and data augmentation, affect REC.  We study the efforts of these  factors from four important aspects of REC, \emph{i.e.,} attribute descriptions, spatial descriptions, the length of expressions and the bounding box accuracy. For attribute descriptions and spatial descriptions, we  run a template parser~\cite{kazemzadeh2014referitgame} to select high-frequency attribute words and spatial words, and calculate the model accuracy on expressions that contain these words. For the length of expressions, we calculate the model performance on expressions of different lengths.  For the bounding box accuracy, we calculate the proportion of the predicted bounding boxes on different IoU metrics.  The statics are given in Fig.~\ref{fig2}.
\begin{figure*}[t]
	\centering
		\vspace{-1em}
	\includegraphics[width=2\columnwidth]{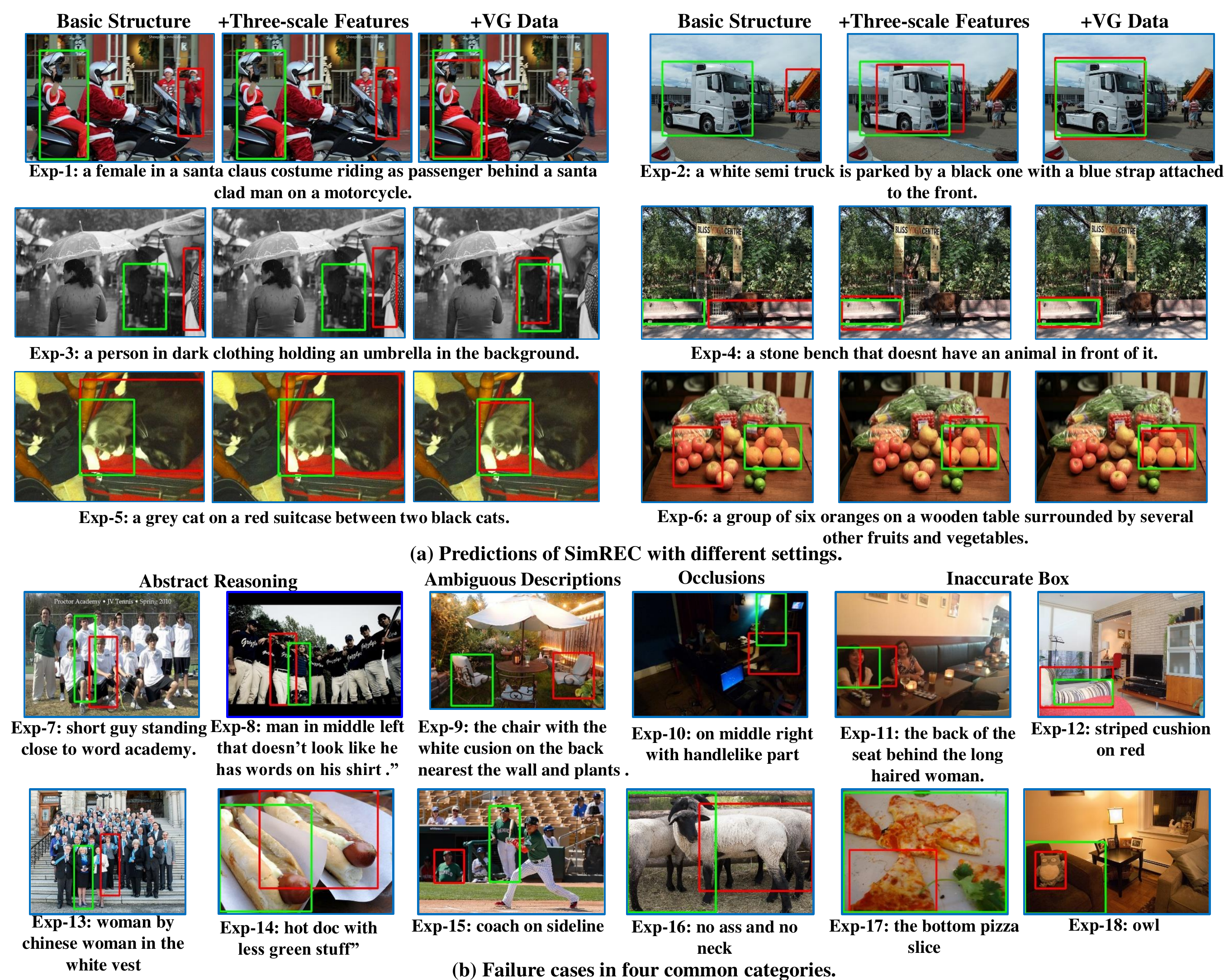}
	%	\vspace{-1em}
	% Reduce the figure size so that it is slightly narrower than the column. Don't use precise values for figure width.This setup will avoid overfull boxes. 
	\caption{Visualizations of predictions by SimREC with different settings  and failure cases. The box of green color is the ground-truth, and the one of red is the prediction. }  
	\label{vis}
	\vspace{-4mm}
\end{figure*}

\subsubsection{Comparison with the state-of-the-art methods}
After making trade-offs between performance and efficiency, we further combine some findings from Tab.~\ref{ablations} to strengthen our baseline network SimREC. The cumulative ablation results are given in Tab.~\ref{cablations}. Then, we compare SimREC to a set of state-of-the-art (SOTA) methods in REC.

%\subsubsection{Cumulative Ablations of SimREC}
%To validate our findings, we conduct detailed cumulative ablations in Tab.~\ref{cablations}. From Tab.~\ref{cablations}, we observe obvious performance gains of each design in SimREC, \emph{e.g.,} up to +29.75\% total improvements on RefCOCO+ \textit{testA} set, which well indicates the effectiveness of findings. Besides, we also notice that the VG pre-training significantly boost the performance, \emph{e.g.,} +6.93\% improvements on RefCOCO+ \textit{val} set. Such a significant improvement  by the pre-training  reveals the great merits of SimREC in  data-efficiency. Conclusively, these results well confirm our findings for SimREC.

%\subsubsection{Tips and Tricks for Robust Performance.}
%\noindent \textbf{Data augmentations}
%\noindent \textbf{Freezing visual backbone}

% \begin{figure}[t]
%	\centering
%	\includegraphics[width=1\columnwidth]{vis1.pdf}
%	
%	% Reduce the figure size so that it is slightly narrower than the column. Don't use precise values for figure width.This setup will avoid overfull boxes. 
%	\caption{Visualizations of predictions by three models. The box of green color is the ground-truth, and the one of red is the prediction. }  
%	\label{vis}
%	\vspace{-2mm}
%\end{figure}

\noindent \textbf{Comparison to one-stage and two-stage SOTAs.} 
We first compare SimREC with   SOTA methods specifically designed for REC, of which results are given in Tab.~\ref{tab:refcoco_results} and Tab.~\ref{SOTA}. For a fair comparison, we remove VG augmentations when compared to these models.  Other training details, such as data augmentation, are also remain  comparable with newly proposed REC models~\cite{zhu2022seqtr,ye2022shifting}.   From this table, we can see that SimREC  outperforms all these methods, \emph{e.g.,} +1.51\%, +2.47\% and +4.01\% on three datasets, respectively. Besides, SimREC also have superior efficiency. With much better performance, SimREC further enhances the speed advantage against two-stage methods, \emph{e.g.,} +21.4 times. Compared with one-stage SOTAs, SimREC is also much more lightweight and efficient,   \emph{i.e.}, {+35.5 fps}. 

{ we also compare SimREC with existing  pre-trained REC models.  Under the similar setups, SimREC achieves better performance than MDETR~\cite{kamath2021mdetr} and UniTAB~\cite{yang2022unitab}, \emph{e.g.,} +4.02\% on RefCOCOg.  Compared to  multi-task approaches~\cite{zhu2022seqtr,yan2023universal,liu2023polyformer}, which 
requires  additional  labeled data,  SimREC also demonstrates comparable results and  significant efficiency.  For instance, SimREC achieves similar performance to PolyFormer-L on RefCOCO  with 22 times faster inference speed. These results greatly confirm that SimREC achieves the better trade-offs between performance and efficiency than existing methods. } 

%As shown in Tab.~\ref{tab:refcoco_results},   SimREC  outperforms  two-stage SOTAs by a large margin, \emph{e.g.,} +6.66\% on RefCOCO testB. In addition, SimREC shows great superiority on inference speed, \emph{e.g.,} 20 times faster than MAttNet~\cite{MATT:}. Compared to one-stage SOTAs, the merits of SimREC are also obvious in two aspects.   Firstly, SimREC consistently outperforms  existing one-stage REC models.  Compared to the transformer-based network TransVG~\cite{deng2021transvg}, SimREC shows distinct improvements on RefCOCO+ by +6.05\%.  Secondly, the simple structure of SimREC makes it more light-weight and faster than existing one-stage REC models.  

%\begin{figure}[t]
%	\centering
%	\includegraphics[width=1\columnwidth]{vis2.pdf}
%	
%	% Reduce the figure size so that it is slightly narrower than the column. Don't use precise values for figure width.This setup will avoid overfull boxes. 
%	\caption{Visualizations of failure cases. The box of green color is the ground-truth, and the one of red is the prediction. }  
%	\label{vis}
%	\vspace{-2mm}
%\end{figure}

\noindent \textbf{Comparison with large-scale BERT-style pre-trained models. }
We  compare SimREC with a set of large-scale BERT-style pre-trained models in Tab.~\ref{tab:pretrained_results}, which includes VilBert~\cite{lu2019vilbert}, ERNIE-ViL~\cite{yu2020ernie}, UNITER~\cite{chen2019uniter} and VILLA~\cite{gan2020large}. 
These methods all apply the expensive BERT-style pre-training with millions of vision-language examples~\cite{krishna2016visual,MSCOCO,sharma2018conceptual,ordonez2011im2text}, and they are also fine-tuned on REC datasets. 
From Tab.~\ref{tab:pretrained_results}, we  surprisingly observe that our simple yet lightweight model can outperform these large models on all benchmark datasets. Compared to VILLA-L~\cite{gan2020large}, the performance gains of SimREC can be up to +11.22\% on RefCOCO \textit{testB}, and the inference time is also reduced by 23 times. Even compared to the recently proposed multi-task model, \emph{i.e.,} OFA~\cite{wang2022ofa}, SimREC also has comparable performance and better efficiency. 
More importantly, the parameter size and training overhead of SimREC are all far less than these methods.

{Since these large models have dominated various V\&L tasks~\cite{VQA1,MSCOCO}, these results actually confirm that one-stage REC can be a ``survivor'' in the era dominated by large BERT-style pre-trained models. From the experiments, we find that one-stage REC is an efficient way for learning vision-and-language alignment. Thus, SimREC can obtain superior performance than the pre-trained models with much fewer training examples.  Meanwhile, the objective of most BERT-style pre-trained models is based on region-text matching and their structures are also modular. In this case, their performance upper-bounds are prone to be limited by the independent visual backbones like Faster RCNN~\cite{ren2017faster}.      }

\subsubsection{Generalization  to Transformer-based REC model}
To validate the generalization ability of SimREC framework, we apply our findings to Transformer-based models, \emph{i.e.,} ViLT~\cite{kim2021vilt} and TransVG~\cite{deng2021transvg}, of which results are given in Tab.~\ref{gen}. From these results, the first observation is that our findings can significantly improve the performance of ViLT~\cite{kim2021vilt} on REC, \emph{e.g.,} +8.94\% on RefCOCO+ testB. Besides, we also find that the performance gains are more obvious on  complex expressions and descriptions about objects. For example, the performances gains are +3.09\% on RefCOCO testA, which are improved to +7.7\% on RefCOCOg test. However, SimREC still outperforms these models on three datasets, greatly confirming the simple  and lightweight structure of SimREC.
Overall, these results greatly validate the generalization ability of our findings on Transformer-based model.

\subsubsection{Qualitative Analysis.}

In Fig.~\ref{vis}, we  visualize the predictions by SimREC with different settings, and also give  some typical failure cases.

From Fig.~\ref{vis}.(a), we can observe that without multi-scale features and data augmentation, SimREC is  easy to fail to understand the expression containing various attributes or relationships. The adoption of multi-scale features and data augmentation can obviously help the model in visual concept understanding as well as relational reasoning. Fig.~\ref{vis}.(b) lists the common failure detections in the well-improved SimREC. {Some of them are attributed to the label noisy, \emph{e.g.}, the 9-\textit{th} and 15-\textit{th} examples, while some are too abstract or beyond the ability of REC. For instance, the 7-\textit{th} example requires the model to understand the characters in the image.   Meanwhile, the object occlusions  also cause the failure of detection, \emph{e.g.,} the 10-\textit{th} example. Besides, we observe that  some failure cases are attributed to the inaccurate regression. For example, SimREC well locates the target object in the 11-\textit{th} and 12-\textit{th} example, but fails in accurately regressing the bounding box of the referent. These examples confirm the effectiveness of our findings, and also demonstrate the upper bound of SimREC.}

\section{Conclusion}
In this paper, we present an empirical study for one-stage REC, which ablates 42 candidate designs/settings via over 100 experimental trails. 
This study not only yields the key factors for one-stage REC in addition to multi-modal fusion, but also reflects some findings  against common impressions about REC. 
By combing the empirical findings, we also improve the simple REC network (SimREC) by a large margin, which greatly outperforms existing REC models in both accuracy and efficiency. More importantly, SimREC  have better performance than a set of large-scale pre-trained V\&L models with much less training overhead and parameters. 
We believe that the findings of this paper can provide useful references for the development of REC, and also give some inspirations for the V\&L research.

% or


\appendices
%% you can choose not to have a title for an appendix
%% if you want by leaving the argument blank
%\section{}

\section*{Acknowledgment}
This work was supported by National Key R\&D Program of China (No.2022ZD0118201), the National Science Fund for Distinguished Young Scholars (No.62025603), the National Natural Science Foundation of China (No. U21B2037, No. U22B2051, No. 62176222, No. 62176223, No. 62176226, No. 62072386, No. 62072387, No. 62072389, No. 62002305 and No. 62272401), and the Natural Science Foundation of Fujian Province of China (No.2021J01002,  No.2022J06001), the China Fundamental Research Funds for the Central Universities (Grant No. 20720220068) and the Major Key Project of PCL (PCL2021A13).


\ifCLASSOPTIONcaptionsoff
  \newpage
\fi



\bibliographystyle{IEEEtran}
\bibliography{IEEEabrv,mybstfile}

%	111
%\end{IEEEbiography}

\begin{IEEEbiography}[{\includegraphics[width=1in,height=1.25in,clip,keepaspectratio]{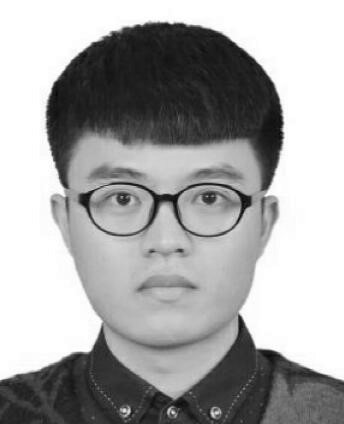}}]{Gen Luo}  is currently pursuing the phd’s
	degree in Xiamen University.
	His research interests include vision-and-language learning.
\end{IEEEbiography}

\begin{IEEEbiography}[{\includegraphics[width=1in,height=1.25in,clip,keepaspectratio]{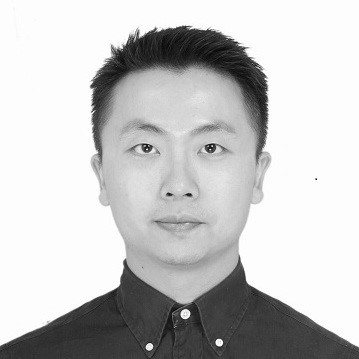}}]{Yiyi Zhou}  received the Ph.D. degree from Xiamen University, China, in 2019, under the supervision of Prof. Rongrong Ji. He was a Postdoctoral Research Search Fellow with Xiamen University from 2019 to 2022. He is currently an associate professor at School of Informatics and Institute of Artificial Intelligence of Xiamen University. His research interests include multimedia analysis and computer vision.
\end{IEEEbiography}

\begin{IEEEbiography}[{\includegraphics[width=1in,height=1.25in,clip,keepaspectratio]{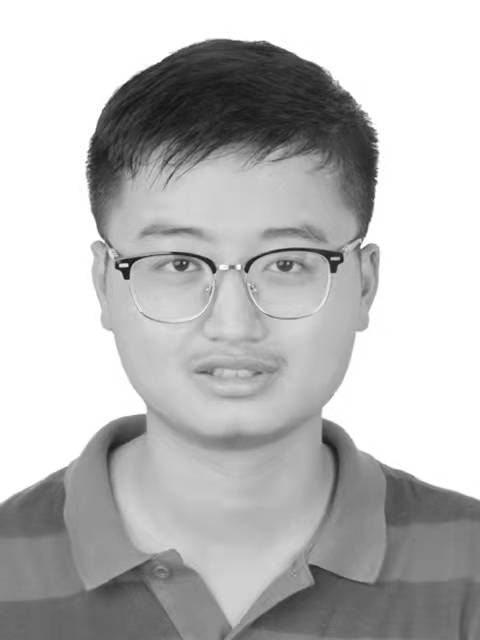}}]{Jiamu Sun}
 received his Bachelor Degree from Hebei University of Technology, China, in 2021. He is a postgraduate student supervised by Prof. Rongrong Ji in Media Analytics and Computing (MAC) lab of Xiamen University, China.
\end{IEEEbiography}

%\begin{IEEEbiography}[{\includegraphics[width=1in,height=1.25in,clip,keepaspectratio]{SBH.jpg}}]{Shubin Huang}
% received his bachelor's degree from FuZhou University, China, in 2021.  He is a postgraduate of the School of Informatics and a member of Media Analytics and Computing (MAC) lab  of Xiamen University, China.
%\end{IEEEbiography}

\begin{IEEEbiography}[{\includegraphics[width=1in,height=1.25in,clip,keepaspectratio]{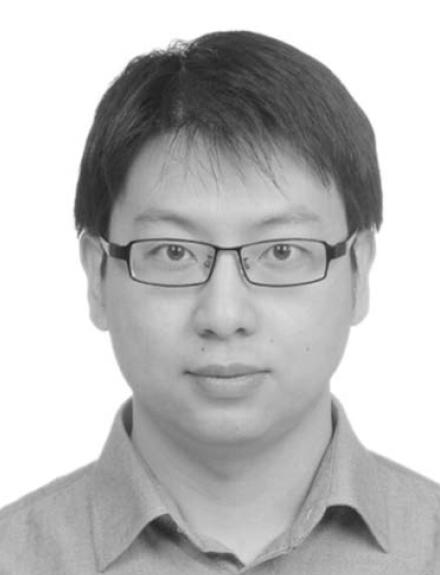}}]{Xiaoshuai Sun} (Senior Member, IEEE)
received the B.S. degree in computer
science from Harbin Engineering University, Harbin,
China, in 2007, and the M.S. and Ph.D. degrees in
computer science and technology from the Harbin
Institute of Technology, Harbin, in 2009 and 2015,
respectively. He was a Postdoctoral Research Fellow
with the University of Queensland from 2015 to
2016. He served as a Lecturer with the Harbin
Institute of Technology from 2016 to 2018. He is
currently an Associate Professor with Xiamen University,
China. He was a recipient of the Microsoft Research Asia Fellowship in 2011.
\end{IEEEbiography}

\begin{IEEEbiography}[{\includegraphics[width=1.1in,height=1.25in,clip,keepaspectratio]{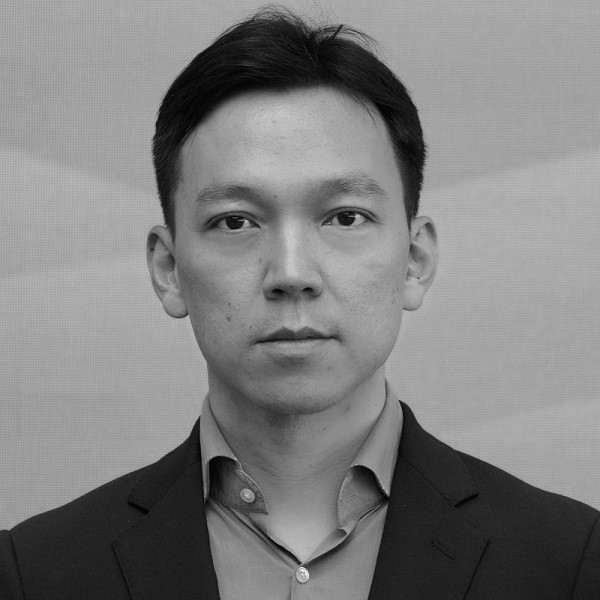}}]{Rongrong Ji}
(Senior Member, IEEE)   is a Nanqiang Distinguished Professor at Xiamen University, the Deputy Director of the Office of Science and Technology at Xiamen University, and the Director of Media Analytics and Computing Lab. He was awarded as the National Science Foundation for Excellent Young Scholars (2014), the National Ten Thousand Plan for Young Top Talents (2017), and the National Science Fundation for Distinguished Young Scholars (2020). His research falls in the field of computer vision, multimedia analysis, and machine learning. He has published 50+ papers in ACM/IEEE Transactions, including TPAMI and IJCV, and 100+ full papers on top-tier conferences, such as CVPR and NeurIPS. His publications have got over 10K citations in Google Scholar. He was the recipient of the Best Paper Award of ACM Multimedia 2011. He has served as Area Chairs in top-tier conferences such as CVPR and ACM Multimedia. He is also an Advisory Member for Artificial Intelligence Construction in the Electronic Information Education Commitee of the National Ministry of Education.
\end{IEEEbiography}


%\newpage



%\vfill


\end{document}